\pdfoutput=1 
\documentclass{article}

\usepackage{amsmath,amsfonts,bm}

\def\eqref#1{equation~\ref{#1}}

\def\1{\bm{1}}

\def\mA{{\bm{A}}}
\def\mB{{\bm{B}}}

\def\mV{{\bm{V}}}
\def\mW{{\bm{W}}}

\def\mZ{{\bm{Z}}}

\DeclareMathAlphabet{\mathsfit}{\encodingdefault}{\sfdefault}{m}{sl}
\SetMathAlphabet{\mathsfit}{bold}{\encodingdefault}{\sfdefault}{bx}{n}
\newcommand{\tens}[1]{\bm{\mathsfit{#1}}}

\def\tZ{{\tens{Z}}}

\usepackage[utf8]{inputenc} %
\usepackage[T1]{fontenc}    %
\usepackage[colorlinks,citecolor=blue!70!black,linkcolor=blue!70!black,
urlcolor=blue!70!black]{hyperref}       %
\usepackage{url}            %
\usepackage{booktabs}       %
\usepackage{amsfonts}       %
\usepackage{nicefrac}       %
\usepackage{microtype}      %
\usepackage{xcolor}         %
\usepackage{wrapfig}
\usepackage{sidecap}
\usepackage{lipsum}
\usepackage{fullpage}
\usepackage{graphicx}
\usepackage{subfigure}
\usepackage{booktabs} %
\usepackage{mkolar_definitions}
\usepackage{enumitem}
\usepackage{comment}

\usepackage{multirow}

\allowdisplaybreaks
\usepackage{natbib}

\usepackage{algorithm}
\usepackage{algorithmic}
\usepackage{xspace}

\usepackage{amsthm,amsmath,amssymb}
\usepackage{amsfonts,dsfont}
\usepackage{mathtools}
\usepackage{tikz}
\usepackage{comment} 
\usepackage{wrapfig}

\usepackage{caption}
\usepackage{parskip}
\usepackage{wrapfig}

\newcommand{\poly}{\texttt{Poly}}
\newcommand{\polys}{\texttt{MHR}\xspace}

\newcommand{\polysmaia}{\texttt{MHR}-Maia\xspace}
\newcommand{\polysrl}{\texttt{MHR}-Rocket\xspace}

\newcommand{\ignore}[1]{{}}

\newcounter{packednmbr}
\newenvironment{packedenumerate}{\begin{list}{\thepackednmbr.}{\usecounter{packednmbr}\setlength{\itemsep}{0.5pt}\addtolength{\labelwidth}{-2pt}\setlength{\leftmargin}{5ex}\setlength{\listparindent}{\parindent}\setlength{\parsep}{1pt}\setlength{\topsep}{0pt}}}{\end{list}}

\title{Generative Modeling of Individual Behavior at Scale}

\author{
  Nabil Omi \qquad Lucas Caccia \qquad Anurag Sarkar \\ Jordan T. Ash \qquad Siddhartha Sen \\
  Microsoft Research\\
}
\date{}

\begin{document}
\maketitle

\begin{abstract}
There has been a growing interest in using AI to model human behavior, particularly in domains where humans interact with this technology. While most existing work models human behavior at an aggregate level, our goal is to model behavior at the individual level. 
Recent approaches to \textit{behavioral stylometry}---or the task of identifying a person from their actions alone---have shown promise in domains like chess, but these approaches are either not scalable (e.g., fine-tune a separate model for each person) or not generative, in that they cannot generate actions. We address these limitations by framing behavioral stylometry as a multi-task learning problem---where each \textit{task} represents a distinct \textit{person}---and use parameter-efficient fine-tuning (PEFT) methods to learn an explicit \textit{style vector} for each person. Style vectors are generative: they selectively activate shared ``skill" parameters to generate actions in the style of each person. They also induce a latent space that we can interpret and manipulate algorithmically. In particular, we develop a general technique for \textit{style steering} that allows us to steer a player's style vector towards a desired property. We apply our approach to two very different games, at unprecedented scales: chess (47,864 players) and Rocket League (2,000 players). We also show generality beyond gaming by applying our method to image generation, where we learn style vectors for 10,177 celebrities and use these vectors to steer their images. 
\end{abstract}

\section{Introduction}

The rapid advances in machine learning in recent years has made it increasingly important to find constructive ways for humans to interact with this technology. Even in domains where AI has achieved superhuman performance, it is often important to understand how humans approach these tasks. Such an understanding can help identify areas for improvement in humans, develop better AI partners or teachers, and create more enjoyable experiences. AI that solely aims to solve a task optimally often fails in these respects, because they tend to be  difficult to interpret, provide limited instructional value, and can be awkward to interact with.

A common method for capturing human behavior is behavioral cloning (BC), a form of imitation learning~\citep{schaal1996learning} that applies supervised learning to fixed demonstrations for a given task.
BC has been used in various domains, such as supply chain management~\citep{supplychain}, legal cases~\citep{adjudication}, robotics~\citep{florence2022implicit}, and self-driving vehicles~\citep{pomerleau1988alvinn}.
Recently, BC has seen increasing use in gaming,
such as in Counter-Strike~\citep{pearce2022counter}, Overcooked~\citep{carroll2019utility}, Minecraft~\citep{schafer-minecraft}, Bleeding Edge~\citep{wham}, and chess~\cite{mcilroy2020maia}. 

The above work focuses on modeling human behavior in aggregate, motivated by goals like developing better AI partners or training tools. However, we believe such goals are better served by modeling human behavior at the \textit{individual} level, 
because this allows us to tailor solutions to the individual's needs (e.g., creating an AI training partner that targets an individual's weaknesses).
To that end, recent work in chess has shown the most promise. 
\citet{mcilroy2020maia}
used BC to create a set of models called Maia that mimic human play at 9 different skill levels. They then fine-tuned these models on the data of 400 individual players to create a personalized model for each player~\citep{mcilroy2022personalized}.
Using these models, the authors perform \textit{behavioral stylometry}, 
where the goal is to identify which person played a given query set of games, by applying each of the 400 models to the query set and outputting the one with the highest accuracy. \citet{mcilroy2021stylometry} propose a more scalable approach of training a Transformer-based embedding on the games of each player. They perform stylometry across 2,844 players by embedding the query set of games and matching it to the closest player's embedding. 

These approaches have different merits. The individualized approach creates a generative model for each player that can play in their style, but it is not scalable: 
adding a new player requires fine-tuning a separate model. The embedding approach is much more scalable: it learns a single-vector representation of each player in a shared style space, and uses few-shot learning to embed a new player in this space. However, it cannot be used to generate moves, and hence cannot reason about player behavior in practice.

An ideal solution would combine these properties: generative, scalable, compact representation. 
Our key insight for achieving this is to view behavioral stylometry as a multi-task learning problem, where each \textit{task} represents an individual \textit{person}. The goal here is to generalize across an initial set of players (tasks) while supporting few-shot learning of new players (tasks). To do this efficiently, we leverage recent advances in parameter-efficient fine-tuning (PEFT)~\citep{ponti2022combining,caccia2022multi}. Specifically, we augment an existing BC model with a set of Low Rank Adapters (LoRAs) as well as a routing matrix that specifies a distribution over these adapters for each player. Unlike approaches that train a separate LoRA for each task, this modular design allows players to softly share parameters in a fine-grained manner. We apply this adapter framework to two very different gaming models: a modified version of the Maia model for chess, and a Transformer-based model for Rocket League, a 3D soccer video game played by cars in a caged arena. We chose these games because they have a large, public collection of human games that span a diversity of skill levels and playing styles.

The base models we create outperform the state-of-the-art BC models for Rocket League and chess.
Our fine-tuning process encourages the adapters to learn different \textit{latent skills}, 
while each row of the routing matrix induces a weight distribution over these skills. We call each row the \textit{style vector} for the corresponding player.
Style vectors are versatile and powerful. They support few-shot learning which enables stylometry at scale. They induce a generative model for each player that we can run and observe. They induce a shared style space that we can interpret and manipulate algorithmically. Leveraging these properties, we develop a general, human-interpretable technique for \textit{style steering} that identifies a subset of players who exhibit a desired style property, and steers a new player towards that property.

This paper makes the following contributions:
\begin{packedenumerate}
\item We use a black-box adapter framework to model individual human behavior and create style vectors for players. We show that style vectors can be combined, interpolated, and steered in an interpretable way.

\item
We perform behavioral stylometry at an unprecedented scale for chess (47,864 players, 94.4\% accuracy) and Rocket League (2,000 players, 86.7\% accuracy), given a query set of 100 games.  
Our per-player generative models achieve move-matching accuracy in the range 45-69\% for chess and 44-72\% for Rocket League.

\item We present novel applications of style vectors, including a method to steer player styles to strengthen human-interpretable attributes of their gameplay. We show the generality of style steering by applying it to image generation for 10,177 celebrities. 

\end{packedenumerate}

\section{Background and Framing}

We frame behavioral stylometry and per-player generative modeling as a multitask learning problem.
In multitask learning \citep{caruana1997multitask, ruder-etal-2019-transfer}, we are given a collection of tasks $\mathcal{T} = \big( \mathcal{T}_1, \ldots, \mathcal{T}_{|\mathcal{T}|} \big) $, where each task $\mathcal{T}_i$ is associated with a dataset $\mathcal{D}_i = \big \{ (x_1, y_1), ..., (x_{n_i}, y_{n_i})\big \}. $ 
Multitask learning exploits the similarities among related tasks by transferring knowledge among them; ideally, this builds representations that are easily adaptable to new tasks.

The premise of this paper is that modeling individual human behavior from a pool of players can be interpreted as a multitask learning problem. In other words, each task $\mathcal{T}_i$ consists of modeling the behavior of a specific player $i$; and dataset $\mathcal{D}_i$ corresponds to the sequence of game actions taken by player $i$, where each $(x,y)$ denotes a game state $x$ 
and the action $y$ that player $i$ took in this state. We use the notion of \textit{tasks} and \textit{players} interchangeably.

\subsection{Parameter-efficient fine-tuning}
Popularized in NLP, parameter-efficient fine-tuning (PEFT) \citep{houlsby2019parameter, lora, tfew} has emerged as a scalable solution for adapting Large Language Models to several downstream tasks. 
Indeed, standard finetuning of pretrained LLMs requires updating (and storing) possibly billions of parameters for each task.
PEFT methods instead freeze the pretrained model and inject a small set of trainable task-specific weights, or ``adapters."

One such approach is the use of Low Rank Adapters (LoRA) \citep{lora}, which modify linear transformations in the network by adding a learnable low-rank shift 
\begin{equation}
\label{eqn:lora}
    h = \big( \mW_{0} + \Delta \mW \big) \ x = \big( \mW_{0} + \mA \mB ^T \big) \ x. 
\end{equation}
Here, $\mW_{0} \in \mathbb{R}^{d \times d}$ are the (frozen) weights of the pre-trained model, and $\mA, \mB \in \mathbb{R}^{d \times r}$ the learnable low-rank parameters of rank $r \ll d$. With this approach, practitioners can trade off parameter efficiency with expressivity by increasing the rank $r$ of the transformation.

\subsection{Polytropon and Multi-Head Adapter Routing}
\label{sec:poly}

Standard PEFT methods such as LoRA can adapt a pretrained model for a given task. In multitask settings, training a separate set of adapters for each task is suboptimal, as it does not enable any sharing of information, or \textit{transfer}, across similar tasks. On the other hand, using the same set of adapters for all tasks risks \textit{negative interference} \citep{wang2021gradient} across dissimilar tasks 
, which may harm optimization and performance. 
Polytropon \citep{ponti2019modeling} (\poly{}) addresses this transfer/interference tradeoff by softly sharing parameters across tasks. 
That is, each \poly{} layer contains 1) an inventory of LoRA adapters 
$$\mathcal{M} = \{\mA^{(1)}\mB^{(1)}, \ \ldots \ , \ \mA^{(m)}\mB^{(m)} \},$$ 
with $m \ll |\mathcal{T}|$, and 2) a task-routing matrix $\mZ \in \mathbb{R}^{|\mathcal{T}| \times m}$, where $\mZ_{\tau} \in \mathbb{R}^{m}$ specifies task $\tau$'s distribution over the shared modules. This formulation allows similar tasks to share adapters, while allowing dissimilar tasks to have non-overlapping parameters. 
The collection of adapters $\mathcal{M}$ can be interpreted as capturing different facets of knowledge, or \textit{latent skills}, of the full multitask distribution.

At each forward pass, \poly{} LoRA adapters for task $\tau$ are constructed as 
\begin{gather*}
\tag{Poly}
\mA^{\tau} =  \sum_{i} \alpha_i \mA^{(i)};\,
\mB^{\tau} = \sum_{i} \alpha_i \mB^{(i)},
\label{eqn:poly}
\end{gather*}
where $\alpha_i = \texttt{softmax}(\mZ\left[\tau\right])_i$ denotes the mixing weight of the $i$-th adapter in the inventory, and $\mA^{(i)},\mB^{(i)},\mA^{\tau}, \mB^{\tau} \in \mathbb{R}^{d \times r}$. 
{Here, the $\tau$-th row of the routing matrix $\mZ$ is effectively selecting which adapter modules to include in the linear combination. 

In our setting, where each task consists of modeling an individual, $\mZ\left[\tau\right]$ specifies which latent skills are activated for user $\tau$; we call this their \textit{style vector}.}
As per Eqn~\ref{eqn:lora}, the final output of the linear mapping modified with a \poly{} LoRA adapter 
becomes 
$ h = \big( \mW_{0} + \mA^{\tau} (\mB^{\tau})^T \big) \ x $. 

In \poly{}, the module combination step remains \textit{coarse}, as only linear combinations of the existing modules can be generated. \cite{caccia2022multi} propose a more fine-grained module combination 
approach, called Multi-Head Routing (\polys{}), which is what we use in our work. Similar to Multi-Head Attention \citep{transfo}, the input dimension of $\mA$ (and output dimensions of $\mB$) are partitioned into $h$ heads, where a \poly{}-style procedure occurs for each head. The resulting parameters from each head are then concatenated, recovering the full input (and output) dimensions. See \ref{app:mhr} for more details.

\paragraph{Routing-only fine-tuning.} \label{sec:z_tuning} While LoRA adapters can reduce the parameter cost from billions to millions \citep{tfew}, training the adapters for each new task can still be prohibitive when dealing with thousands of tasks. To this end, \cite{caccia2022multi} proposed routing-only finetuning,
where after an initial phase of pretraining, the adapter modules are fixed, and only the routing parameters $\mZ$ are learned for a new task. This reduces the parameter cost for each additional task by several orders of magnitude, while maintaining similar performance.
We use this method for few-shot learning.

\section{ML Methodology}

\begin{figure}[t]
\vspace{-10pt}
           \centering
           \includegraphics[width=1.0\linewidth]
           {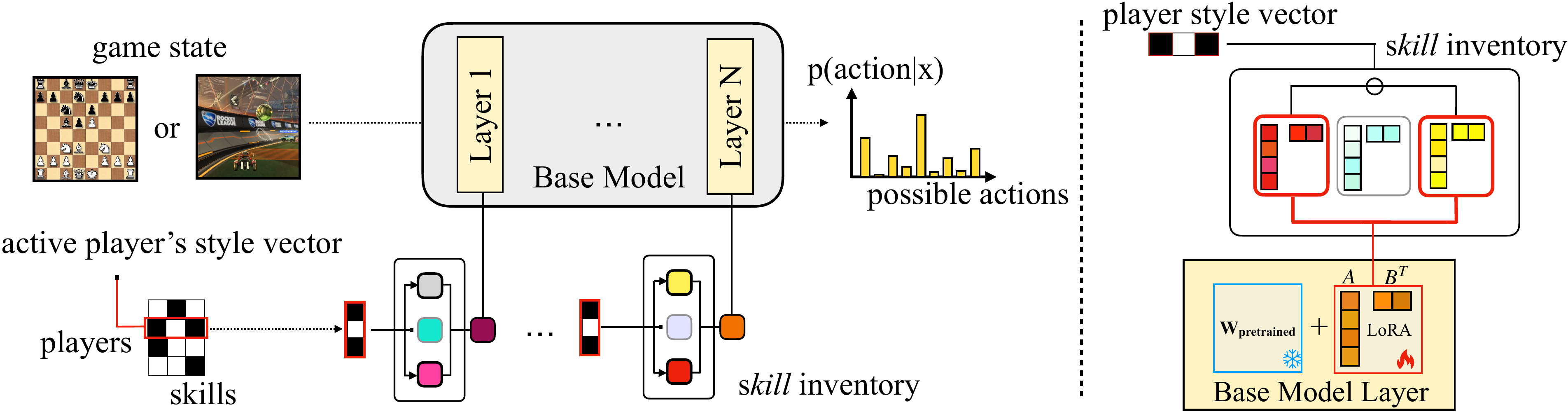}
        \caption{(left) Our overall architecture. We augment a base model with a set of MHR adapters and a routing matrix composed of each player's style vector. (right) Detailed view of an MHR layer, showing a skill inventory of adapters shared across players. The player's style vector specifies which skills are active (in this case, the first and third) to generate the final low-rank weight shift that is applied to the (frozen) base model layer.\looseness=-1}
\label{fig:setup}
\end{figure}

We detail our methodology for creating a generative model of individual behavior that enables our style analyses. 
We start with a base model and apply the \polys adapter framework to it, and then discuss model training and evaluation.

\subsection{Model architecture}

For chess, we follow \cite{mcilroy2022personalized} and use the Squeeze-and-Excitation (S\&E) Residual Network~\citep{hu2018squeeze} as a base model, but with a deeper and wider configuration (see ~\ref{app:maia-details}). At every residual block, an additional 2-layer MLP rescales the residual output along the channel dimension to explicitly model channel  interdependencies.
The input is a 112-channel 8$\times$8 image representation of the chess board; the output is the predicted move encoded as a 1858-dimensional one-hot vector. The total parameters is 15.7M.  
For Rocket League, we use the GPT-2 architecture from \cite{radford2019language} with a dimensionality of 768, 12 attention heads, and 12 layers. The input is a 49-dimensional vector with game physics information; the output is 8 heads: 5 with 3 bins of [-1, 0, 1] and 3 binary heads for a total of 1944 possible action combinations. The model has no embedding layer, as the game data points are passed directly as tokens after processing. The total parameters is 87.7M. 

To enable user-based adaptation, we incorporate the \polys{} adapters described in Section~\ref{sec:poly} into our base models, as illustrated in Figure~\ref{fig:setup}. In chess, for every linear transformation in the MLP used for channel-wise rescaling, we add an \polys{} layer built of LoRA adapters with rank 16, for a total of 12$\times$2~=~24 \polys{} layers. We use an adapter inventory of size 32 and a multi-head routing strategy with 8 heads. Therefore, for each user we must learn 32$\times$8~=~256 routing parameters as their style vector. This yields 5M additional parameters.  For Rocket League, we attach the adapters to the fully connected layer of each transformer block, resulting in 12 \polys{} layers of LoRAs with rank 16. We use an inventory size of 16 and 64 heads. This yields 13.8M additional parameters. To facilitate interpretability and style analysis, we use the same routing (style vector) across all \polys{} layers.\looseness=-1

\subsection{Data collection and partitioning}
We use data from the largest open-source online chess platform, Lichess.org~\citep{lichess}, which boasts 
a database of over 6 billion games. 
We collected Blitz games played between 2013 and 2020 inclusive---these are games with 3 or 5 minutes per side, optionally with a few seconds of time increment per move---and applied the same player filtering criteria as~\cite{mcilroy2022personalized}. 
The resulting dataset comprises 47,864 unique players and over 244 million games. 
(See \ref{app:maia-details} for a discussion on data imbalance.) 
For Rocket League, we collect data from a large open-source replay database, Ballchasing.com~\citep{ballchasing}. %
We use 2.2 million 1v1 replays from 2015 to mid-2022, totalling several decades of human game play hours at 5 minutes per game. 
After parsing, each Rocket League game state is a vector holding the player's 3D position, linear and angular velocity, boost remaining, rotation, and team; we also include the opponent's state and the position, linear and angular velocity of the ball. Given a game state, we have to predict the user's throttle, steer (while grounded), pitch, yaw, roll (while aerial), jump, boost, and handbrake. 
Additional logic was needed to correct for missing aerial controls and inconsistent sampling rates (24-27hz). We detail our data processing procedure and challenges in \ref{app:rocketleaguedata}.

We divide the set of players into a few subsets to support our training methodology. The \textit{base player} set comprises all data and is used to train the base models. 
The \textit{fine-tuning player} set is used to fine-tune the \polys architecture shown in Figure~\ref{fig:setup}. (For both, we split each player's data into 80/10/10 for train/test/validation.) The \textit{few-shot player} set is used for few-shot learning based on a reference set of 100 games per player. 
For our chess experiments, to enable a direct comparison with prior work,
we create an additional fine-tuning player set consisting of the same 400 players from those studies. Currently, we treat each player's data holistically, but in principle one could partition a player in different ways to analyze their playing style (as discussed in \ref{app:assumptions}).

\subsection{Model training and evaluation}
\label{subsec:traineval}

\paragraph{Base model.}
We train our base Maia model for chess using data from a base player set of all 47,864 players, treating this as a classification task of predicting human move $y$ made in chess position $x$, given a datapoint $(x,y)$. 
We use the same loss functions and evaluation criteria as the original Maia work: Maia's policy head uses a cross entropy loss while the value head uses MSE; the output of the policy head is used to evaluate the model's move-matching accuracy. 

We train our Rocket League model using a base player set of over 800,000 players, though the vast majority of players have 5 games or fewer. 
We %
discretize the actions into 3 bins for throttle, steer, pitch, yaw, and roll, as most of this data is close to 0, -1, or 1. We use binary outputs for jump, boost, and handbrake. 
A next-move prediction is labelled correct if and only if all the outputs are correct.

\paragraph{\polys fine-tuning.}
To train the \polys LoRA adapters, 
we adopt the methodology used in~\cite{caccia2022multi}: namely, we {freeze the base model} and fine-tune the \polys layers and routing matrix using data from a fine-tuning player set. Recall that the routing matrix $\mZ$ has a row (style vector) for each player in the fine-tuning set. 
Following \cite{ponti2019modeling}, we use a two-speed learning rate, where the style vector learning rate is higher than the adapter learning rate.

For chess, we use two fine-tuning player sets in our experiments, creating two separate \polysmaia models. The first set comprises all 47,864 players and is used to evaluate behavioral cloning and stylometry at very large scale. The second set is comprised of the same 400 players used by~\cite{mcilroy2022personalized}, which we use to compare few-shot learning and stylometry results. 
For Rocket League, we train an \polysrl model on a fine-tuning set of 2,000 players with 100 games each.

\paragraph{Few-shot learning.} 
To perform few-shot learning on our \polys{} models, we perform the ``routing-only fine-tuning" described in section \ref{sec:z_tuning} that additionally freezes all \polys LoRA adapters. Given a few-shot player, we add a (randomly-initialized) new row to $\mZ$ and fine-tune it on the player's reference set of games, eventually learning a style vector for the player.
Using this style vector, we can invoke a generative model of the player and use it to evaluate move-matching accuracy, as described above. To perform stylometry, if the player is a \textit{seen} player (i.e., part of the fine-tuning set), then a matching style vector already exists in $\mZ$, and we can find it using cosine similarity. Otherwise, if the player is \textit{unseen}, then we simply repeat the few-shot learning process on a query set of games (from the same player), and compare this new style vector to the entries in $\mZ$. In general, the number of reference/query games required for few-shot learning is low (see Figure~\ref{fig:cosine_similarity_vargame} in~\ref{app:maia-details}).

For chess, (unless stated otherwise), all of our few-shot experiments use the \polysmaia model fine-tuned on the 400-player set from~\cite{mcilroy2022personalized}. For Rocket League, the few-shot player set consists of 1,000 of the 2,000-player set used to fine-tune \polysrl. 

\paragraph{Evaluation.}
We evaluate a fine-tuned \polys model in two ways. 
First, we measure its move-matching accuracy, similar to how we evaluate the base models. However, since our \polys models provide a generative model for each player (conditioned on their style vector), we can separately evaluate each player's model by applying it to their test set.
We then average these per-player accuracies to determine the overall move-matching accuracy for the model.

Our second evaluation method uses the model to perform behavioral stylometry among all players in the fine-tuning set. In theory, we could adopt the methodology of~\cite{mcilroy2022personalized} and compute the move-matching accuracy of every player applied to every other player's query set, but such a quadratic computation is infeasible beyond a few thousand players. 
Instead, we leverage our few-shot learning methodology above.
That is, given a query set of games from some player, we learn a new style vector in $\mZ$ for those games via  few-shot learning, and compare this vector to every other vector in $\mZ$ using cosine similarity. We then output the player with the highest cosine similarity. In domains that focus on authenticating individuals (e.g., biometrics), ROC curves and related metrics are used. Our results can be re-interpreted in this way; Figure~\ref{fig:roc_curve} in the appendix shows an example.

\section{Style Methodology}
\label{sec:style-methodology}

The style vectors in $\mZ$ 
give us a starting point for comparing player styles. For example, our stylometry method above uses the cosine similarity between vectors to determine how similar or different players are. 

Style vectors can also be learned for different partitions of a player's dataset, or even for a merged dataset comprising multiple players. The latter is notable because it actually creates a new (human-like) playing style that has never been seen before. This suggests a general approach to synthesizing new styles: interpolate between existing players using a convex combination of their style vectors. To determine the playing strength of a newly synthesized player, we can simulate games between them and the players they are derived from, by conditioning the \polys model on their respective style vectors. The results of these games yield a win rate, which can be converted to a strength rating. 

The latter method is notable because it creates a new playing style that is human-like, and yet has never been seen in the world. This suggests a more general approach to synthesizing new styles: interpolate between existing players using a convex combination of their style vectors. 
For example, we can smoothly transition from a weaker player's style to a stronger player's style. 

Currently, our advanced style synthesis techniques focus on chess, where simulating games is cheap and evaluation heuristics are standardized.
Rocket League simulations are too costly at present for this, and there are no standardized heuristics, but in principle the same methodology can be applied. We plan to explore this in future work.

In order to make style comparisons more human interpretable, 
we draw inspiration from the concept probing technique used to analyze AlphaZero (a deep reinforcement learning chess engine)~\citep{acquisition-alphazero}. We use a set of human-coded heuristic functions found in Stockfish (a traditional chess engine) 
to evaluate a player's model.  
These functions capture concepts such as: king danger, bishop pair utilization, material imbalance, and so on. By invoking a player's model on a fixed set of chess positions, we can measure the change in the heuristic functions before and after their chosen move, and 
use this to summarize how much emphasis the player places on the corresponding concept.

Combining the above methods, we propose a simple but general method for \textit{steering} a player's style towards a specific, human-interpretable attribute $a$ (e.g., king danger), while limiting the changes to other attributes (so as to preserve their style). We summarize this method in Algorithm \ref{alg:style_vector}. 
We first collect a set of players $X$ who exhibit high values for attribute $a$---determined, for example, by running their generative models on a fixed set of game states. We extract the common direction among these players, by averaging their style vectors and subtracting the population average. This yields a \textit{style delta vector} that can be added to any player's style vector to elicit the desired change. 

\begin{algorithm}[H]
\setlength{\algorithmicindent}{-10pt}    
\begin{algorithmic}  
   \STATE {\bfseries \hspace{-4pt} Input: } \\
   $X$ : Style vectors of top-k players for attrib. $a$; \\ 
   $P$ : Style vectors of all players in population  \\   
   \STATE {\bfseries \hspace{-4pt} Output} $\Delta_a$: \text{Style delta vector for attr. }$a$  \\  
   \vspace{5pt}  
   \STATE $\mV_{a} =  \text{mean}(X, \texttt{axis}=\text{`players'})$  \\  
\STATE $\mV_{P} =  \text{mean}(P, \texttt{axis}=\text{`players'})$  \\  
 
  \STATE $\Delta_{a} = \mV_{a} - \mV_{P}$
   \STATE \textbf{Returns} $\Delta_{a}$ \\  
\end{algorithmic}  
\caption{  
Style Delta Vector computation}  
\label{alg:style_vector}  
\end{algorithm} 

\section{Experiments}

In this section, we show that \polysmaia performs competitively with prior methods for behavior cloning and stylometry in chess, and does so at an unprecedented scale. 
We also apply our approach can be applied to Rocket League, a challenging 3D video game environment, for both stylometry and move prediction. Next, we show that explicitly capturing style vectors allows us to analyze and manipulate the behavior of player models.  Finally, demonstrate that our approach generalizes beyond gaming, to personalized and steerable image generation.

\begin{table}
\small
    \centering
    \begin{tabular}{lllcrr}
        \toprule
         Method & $|\textit{Query}|$ &  $|\textit{Universe}|$ &  $|\textit{Query Games}|$ &  Random (\%)  &  Acc. (\%)  \\
        \midrule
        \vspace{2pt}
            \cite{mcilroy2022personalized}  &    400 &    400 &  100 &    0.25 & 98.0 \\
            \cite{mcilroy2021stylometry}    &    400 &    400 &  100 &    0.25 & 99.5 \\
        \vspace{3pt}
           \polysmaia                            &    400 &    400 &  100 &    0.25 & \textbf{99.8} \\
           \midrule
            
            \cite{mcilroy2022personalized}  &    400 &    400 &  30 &    0.25 & 94.0 \\
             \vspace{3pt}
            \polysmaia                            &    400 &    400 &  30 &    0.25 & \textbf{98.8} \\
            \midrule

           \polysmaia (seen)                            &  10000 &  47864 &  100 &    0.002 & 94.4 \\  
        \polysmaia (unseen few-shot)      &    10000 &    10000 &  100 &    0.01 & 87.6 \\
        \bottomrule
    \end{tabular}
    \caption{Top-1 stylometry accuracy results. Random (\%) indicates the chance of choosing the correct player when randomly sampling, and defines how difficult the task is.
    Numbers for \cite{mcilroy2022personalized} and \cite{mcilroy2021stylometry} are borrowed from their respective papers; the same 400 player dataset is used across all comparisons. When increasing the universe size from 400 to 47,864 players, \polysmaia's accuracy drops by only 5.4\%, demonstrating its scalability. }
    \label{tab:stylometry_known_users}

\end{table}

\subsection{Behavioral Stylometry}

For chess, we show that \polysmaia perform competitively with previous behavioral stylometry methods for both seen and unseen players. Here, the goal is to predict which player produced a given set of games. We compare our approach to individual model fine-tuning \citep{mcilroy2022personalized}, which fits a separate pre-trained Maia model to the data of each player, and to a Transformer-based embedding method \citep{mcilroy2021stylometry}, which embeds players in a 512-dimensional style space based on their games. All reported accuracies are top-1 unless stated otherwise.

To perform stylometry on a query set of games, \citet{mcilroy2022personalized} apply each player's fine-tuned Maia model on the query set and select the one with the highest move-matching accuracy. 
As seen in Table~\ref{tab:stylometry_known_users}, this procedure works well, but it is very expensive---requiring a separate model for each player as well as computationally intensive inference calls on the entire query set per player.

In contrast, both the Transformer-based embedding and \polysmaia scale to much larger numbers of players. The Transformer-based embedding needs only to embed the query games to compute a player vector, while \polysmaia needs only to fit a new style vector on the query games. In either case, the produced vectors are compared to those in the player set to find the closest match, e.g., using cosine similarity.
Table~\ref{tab:stylometry_known_users} compares these approaches, showing that \polysmaia performs competitively, and scales well to large universe sizes.
When performing stylometry on \textit{seen} players,
we sample 10,000 query players from the set of seen players and fit a new style vector for each query player based on their 100-game query set. 
For stylometry on \textit{unseen few-shot} players, we start with an \polysmaia model trained on the 400 player dataset from ~\cite{mcilroy2022personalized}, sample 10,000 players from a held-out set, fit style vectors based on their 100-game reference sets, and then apply the above methodology on their query sets. We achieve an accuracy of 87.6\% with this method. In comparison, \cite{mcilroy2021stylometry} achieves 79.1\% using the same 400 player training dataset and a smaller (different) universe of 2,844 players.  Although the Transformer-based embedding method can scale similarly in training and inference to our method, it is not a generative model (i.e., cannot play the game), which severely limits its utility.

For Rocket League, to the best of our knowledge, we are the first to attempt stylometry. We apply the same few-shot learning methodology to compute style vectors for 1,000 query players based on their 100-game reference sets, and then fit a new style vector for each query player based on their 100-game query set. (Recall that each game consists of 5 min of 1v1 gameplay.) For each of the 1,000 query players, our \polysrl approach must correctly identify them among a universe of 2,000 players. We achieve an accuracy of \textbf{86.7\%}, 
showcasing the validity of our approach even in a challenging 3D game scenario.

\looseness=-1

\subsection{Move generation}

\begin{wrapfigure}[19]{R}{0.42\textwidth}
  \centering
  \includegraphics[width=0.42\textwidth]{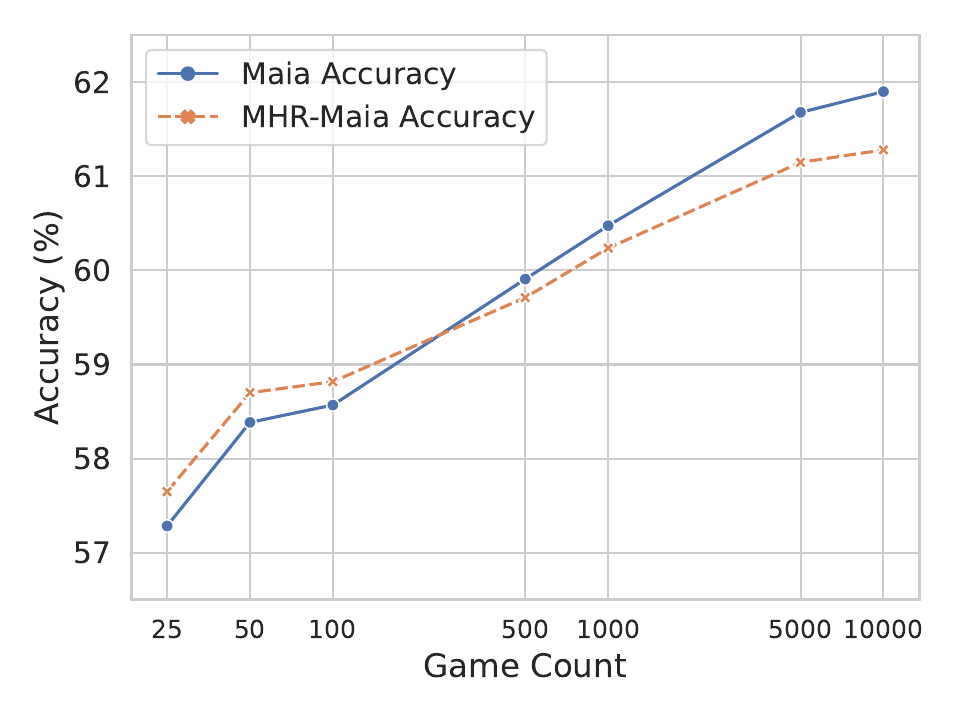}
  \vspace{-20pt}
  \caption{Accuracy at various game counts of the individual models (Maia) and our method (\polysmaia). \polysmaia is within 1\% accuracy of individual model fine-tuning using roughly 1\% of the compute cost per player.}
  \label{fig:maia_poly_comparison}
\end{wrapfigure}

A key feature of our \polys models is that they are generative.
For chess, we compare the efficacy of our method to using individually fine-tuned models for each player. We do not compare to the Transformer-based embedding method because it is incapable of generating moves. Full fine-tuning of individual models generally results in superior performance compared to PEFT methods, albeit at much higher computational cost.
Nevertheless, Figure~\ref{fig:maia_poly_comparison} shows that \polysmaia performs comparatively well, 
achieving within 1\% accuracy of individual model fine-tuning over a wide range of game counts. We achieve this using roughly 1\% of the compute cost per player, as follows. On an A100 80GB GPU, training individual models required roughly 20 A100-minutes per player on average; thus, training on the full 47,864 player dataset would require thousands of A100-hours. In contrast, training \polysmaia on the full dataset required roughly 7 A100-days, or around 12-13 A100-seconds per player, an improvement of nearly two orders of magnitude. The inference costs were roughly equal, with \polysmaia being marginally more expensive due to the added parameters.
The 47,864 player \polysmaia model achieves a mean move prediction accuracy of 59.0\%, while our base model achieves 54.4\%. The per-player accuracies range from 45-69\%.

For Rocket League, we compare the next move prediction of our base model (trained on over 800,000 players) with \polysrl (fine-tuned on 2,000 players), to show that player-based conditioning via style vectors generates better predictions. 
We find that \polysrl increases the next move prediction accuracy from \textbf{53.1\%} to \textbf{56.1\%} (random performance being 0.05\%), averaged over all 2,000 players.
Moreover, the per-player move-matching accuracy ranges from 44-72\%, suggesting that the model has learned a wide range of non-trivial behaviors.

\subsection{Analysis of style vectors}

In this section, we explore the consistency of our style vectors within a player and across different players. 
We also compare playing styles using human-interpretable metrics, and generate new styles by averaging style vectors.

\begin{figure}[tbp]  
    \centering  
    \begin{minipage}{\textwidth}  
        \centering 
        \includegraphics[width=0.7\textwidth]{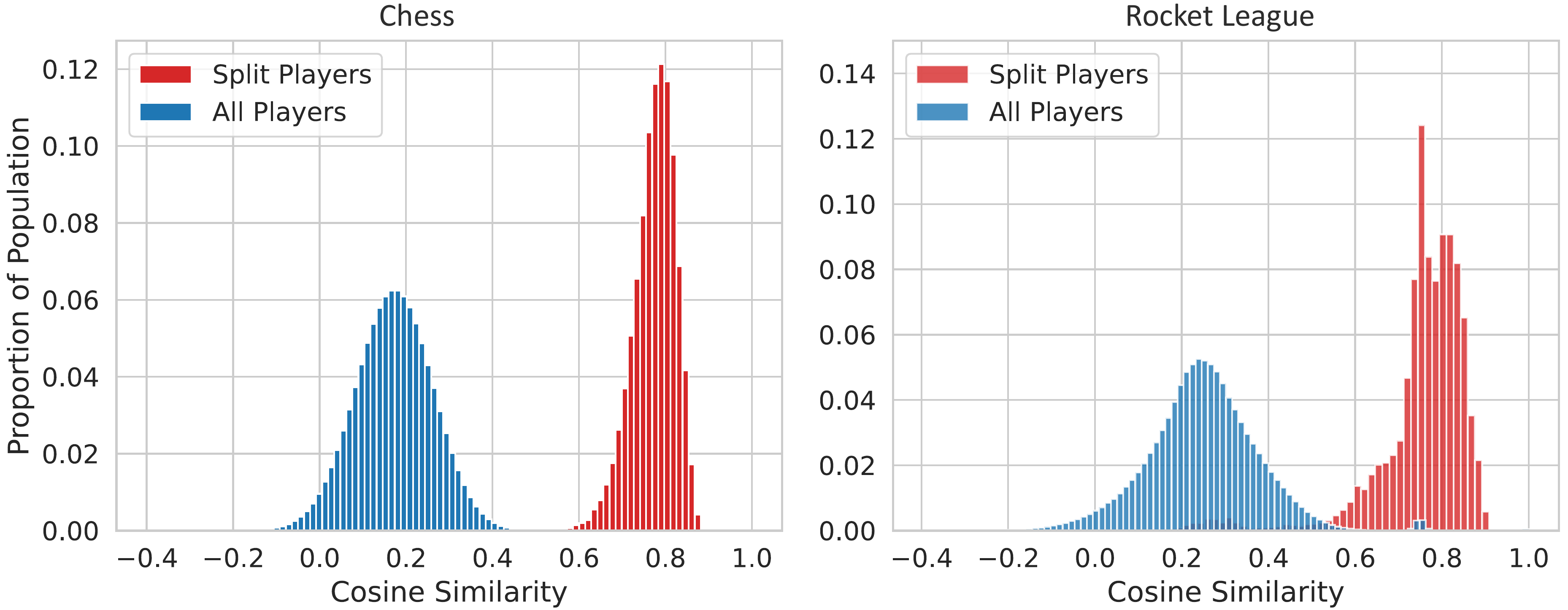}  
        \caption{The distribution over cosine similarity between style vectors learned from different partitions of the same player (red) vs across all players (blue). A pair of style vectors learned from non-overlapping portions of a single player's data are far more similar to each other than those learned from distinct players.}
        \label{fig:split_cosine}  
    \end{minipage}  

\end{figure}

\begin{figure}[t]
\centering
    \begin{minipage}{0.44\textwidth}  
        \centering  
        \includegraphics[width=0.7\textwidth]{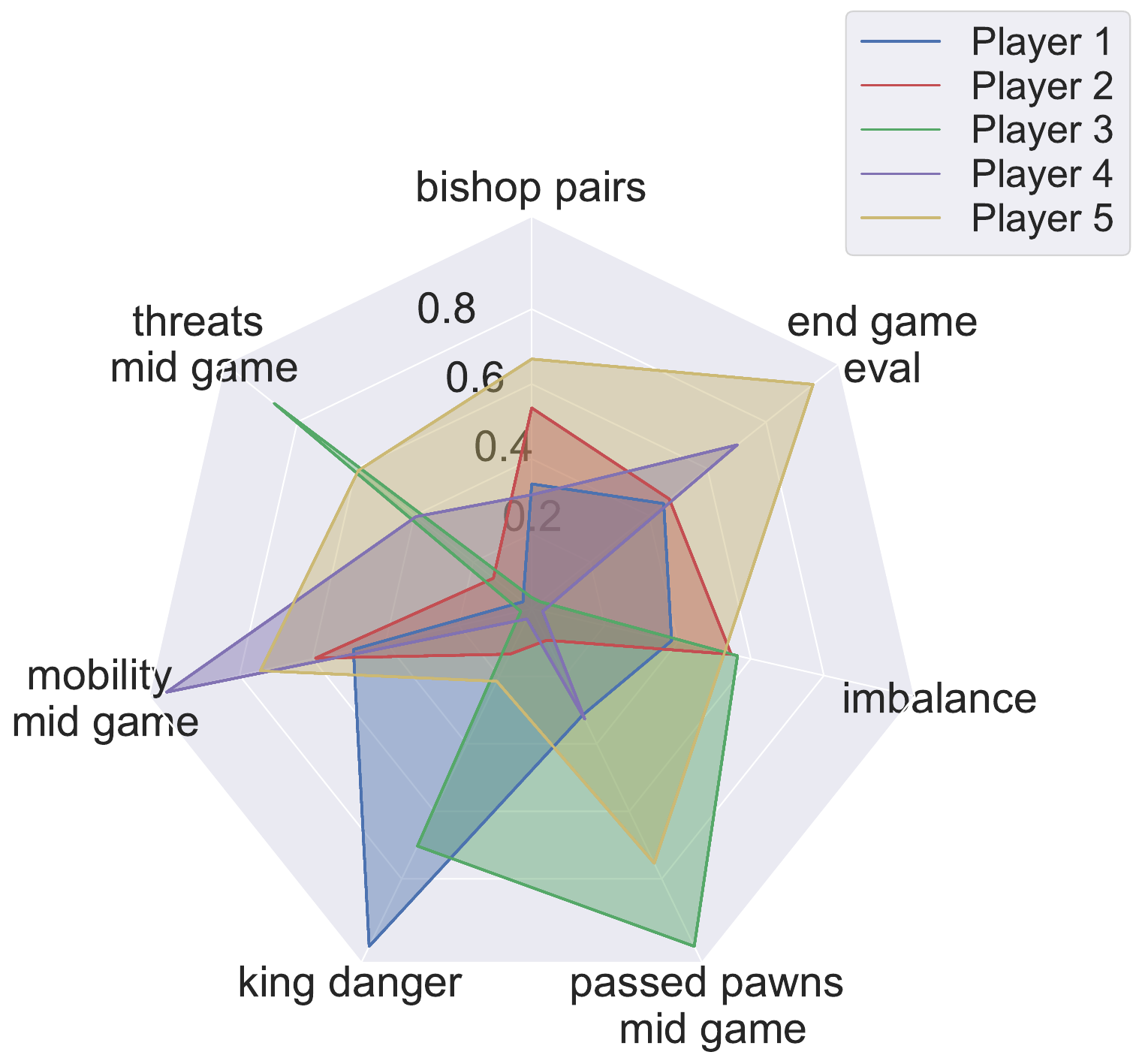}  
        \caption{Comparing different player styles using human-interpretable evaluation metrics.}  

        \label{fig:spiderplot}  
    \end{minipage}  
    \hspace{0.2in}
    \begin{minipage}{0.44\textwidth}
    \centering
    \includegraphics[width=0.7\textwidth]{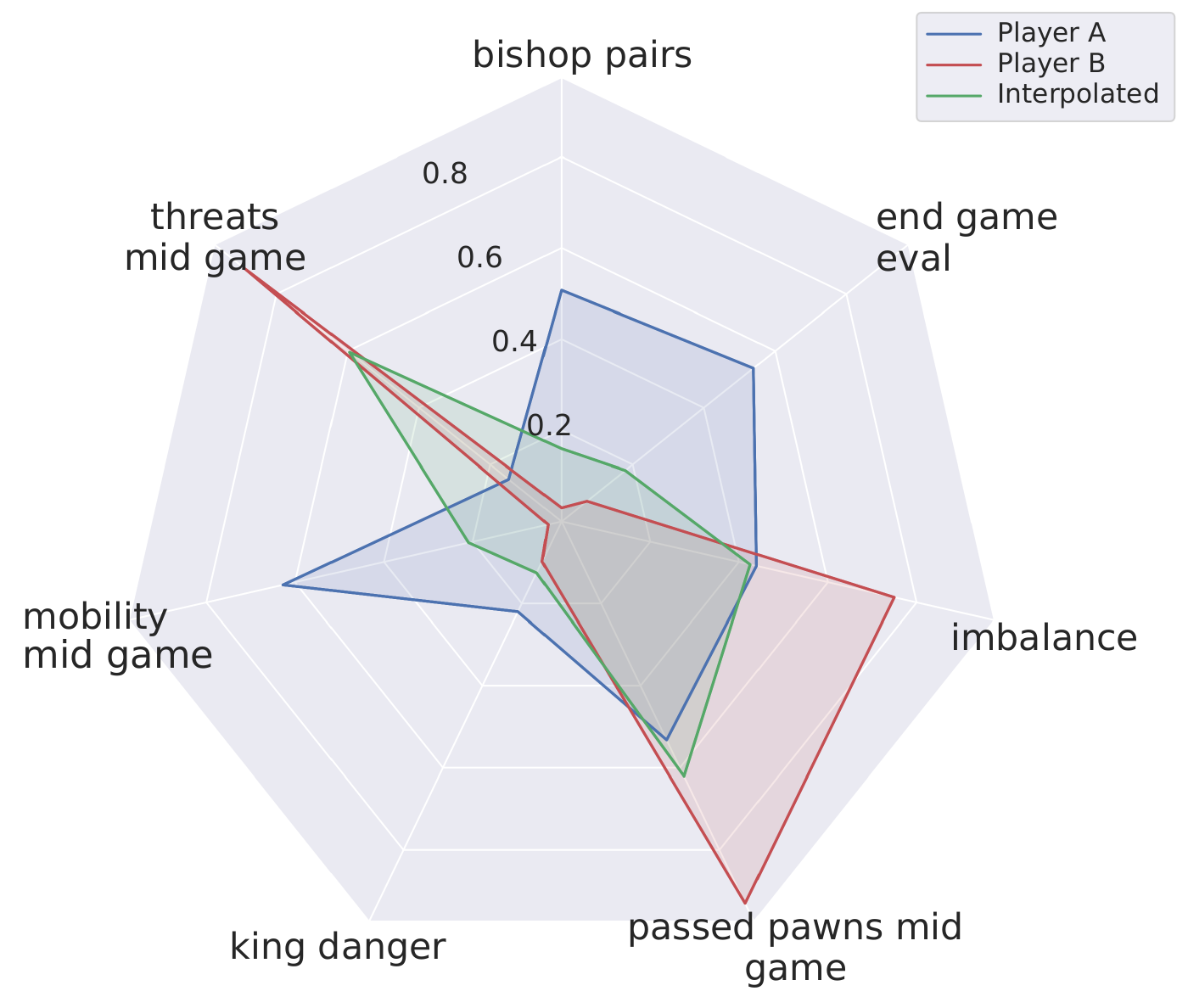} 
    \caption{The style of an intermediate player (green) is shown along with the two component players (blue and red).}
    \label{fig:merged_spider}  
    \end{minipage}
\end{figure}

\paragraph{Consistency within a single player.} To investigate if style vectors show consistency within a player, we first partition 50 players' datasets into disjoint subsets. We use 50 splits for chess and 20 for Rocket League. The subsets are sampled across a wide range of dates, opposing players, and playing sessions. We then train a style vector on every split, and compare these vectors using cosine similarity. We find that vectors corresponding to the same player are significantly more similar to each other than the general population,
visualized in Figure~\ref{fig:split_cosine}. This suggests that our \polys models are able to associate distinct style characteristics with each player.
These characteristics can be learned with relatively few games, as shown in Figure~\ref{fig:cosine_similarity_vargame}.
Additionally, these style characteristics are diverse: we sampled 5 random chess players and used their models to predict their preferred move across $2^{17}$ positions, and then evaluated the move choices 
according to the Stockfish heuristics from Section \ref{sec:style-methodology}. We present the averaged metrics for each player in Figure~\ref{fig:spiderplot}, demonstrating that style vectors indeed capture a wide diversity of playing styles.

\begin{figure}[t]
\centering
    \hspace{-0.3in}\includegraphics[width=0.7\textwidth]{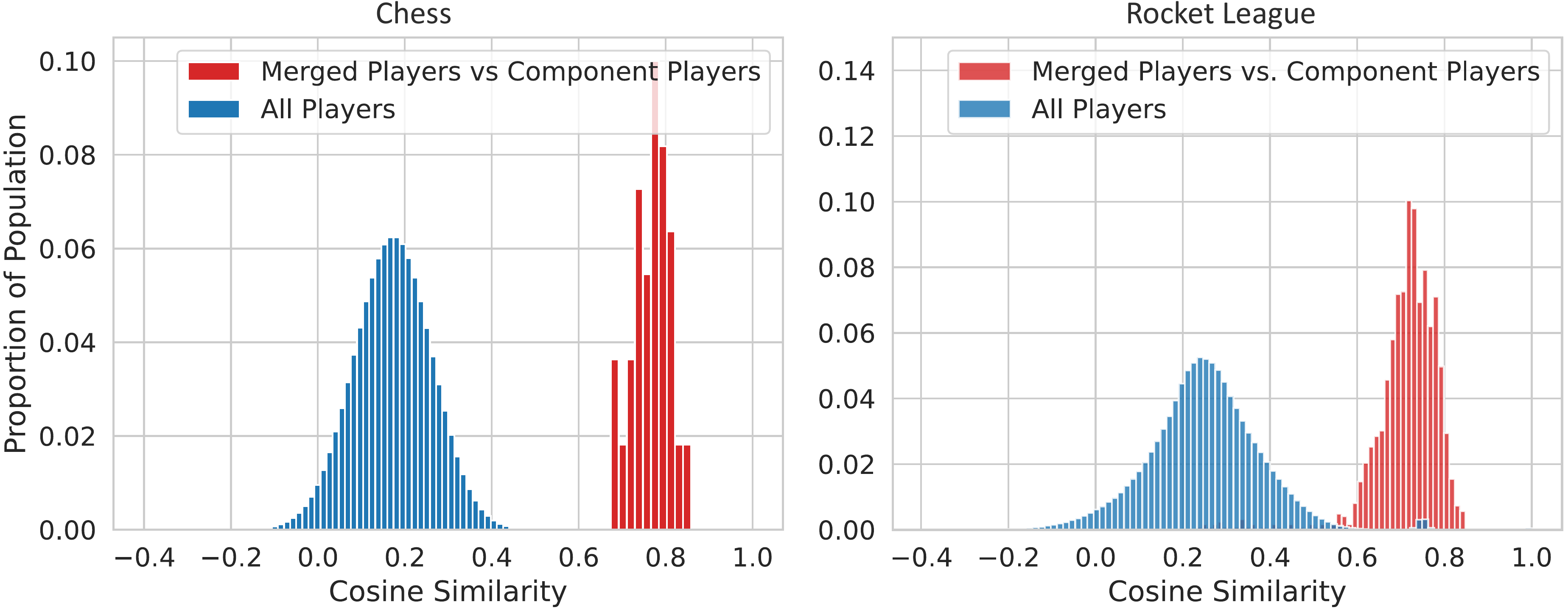}
    \caption{Cosine similarity between averaged style vectors of two players, and the learned style vectors on their merged datasets (red) vs across the full population (blue).}
    \label{fig:merged_cosine}
\end{figure}

\paragraph{Consistency across merged players.}
To investigate if style vectors show consistency across different players, we merge two players' datasets to create a new dataset representing the characteristics of both players. We train a style vector on the merged dataset, and then compare this to the vector obtained by simply averaging the style vectors of the two players. Figure~\ref{fig:merged_cosine} shows the results across a large population of players in Chess and Rocket League. The style vectors trained on the merged datasets have high cosine similarity with the averaged vectors of the component players, while having low similarity with the general population in both games. As an example in chess, we sampled two players and averaged their style vectors, then evaluated the new player's moves across 4096 games using the Stockfish heuristics. This is visualized in Figure~\ref{fig:merged_spider}, showing that the style characteristics of the new player (green) intermediate between the styles of the component players (red, blue).

\subsection{Synthesis of new styles}
\label{sec:synthesis}

In this section, we investigate more advanced applications of style synthesis that can help humans improve: interpolating weaker player styles to stronger ones, and steering player styles along human-interpretable properties.

\begin{figure}[t]  
    \centering  
    \begin{minipage}{0.45\textwidth}  
        \centering  
        \includegraphics[height=0.65\textwidth]{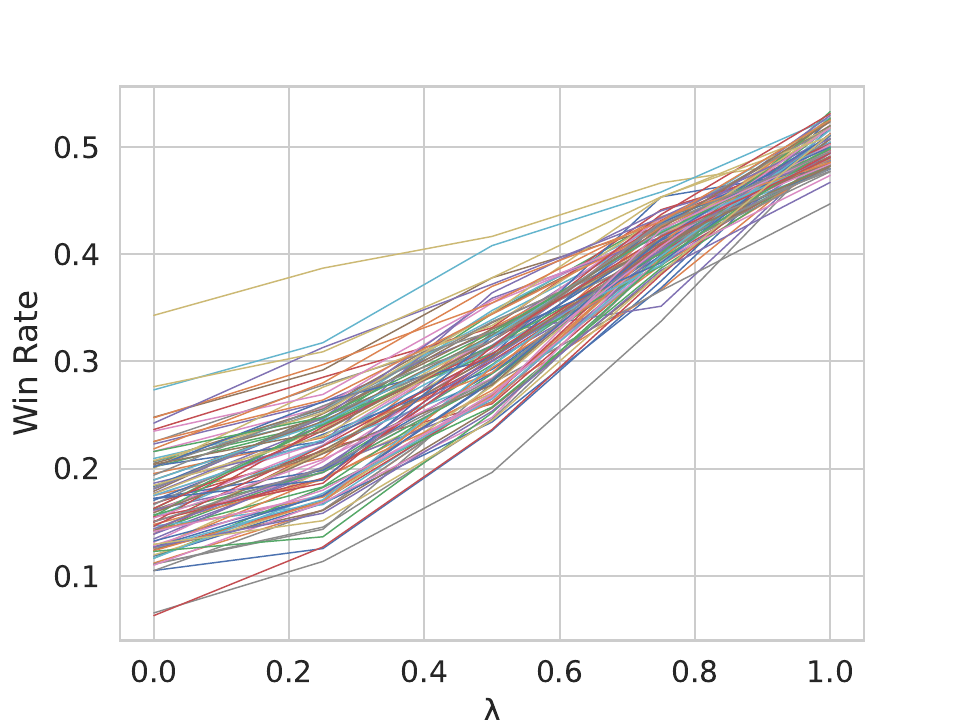}  
        \caption{Win rate as randomly chosen weaker players are interpolated towards randomly chosen stronger players.}  
        \label{fig:interpolation}  
    \end{minipage}%
    \hspace{0.1in}
    \begin{minipage}{0.45\textwidth}  
        \centering  
        \includegraphics[height=0.6\textwidth]{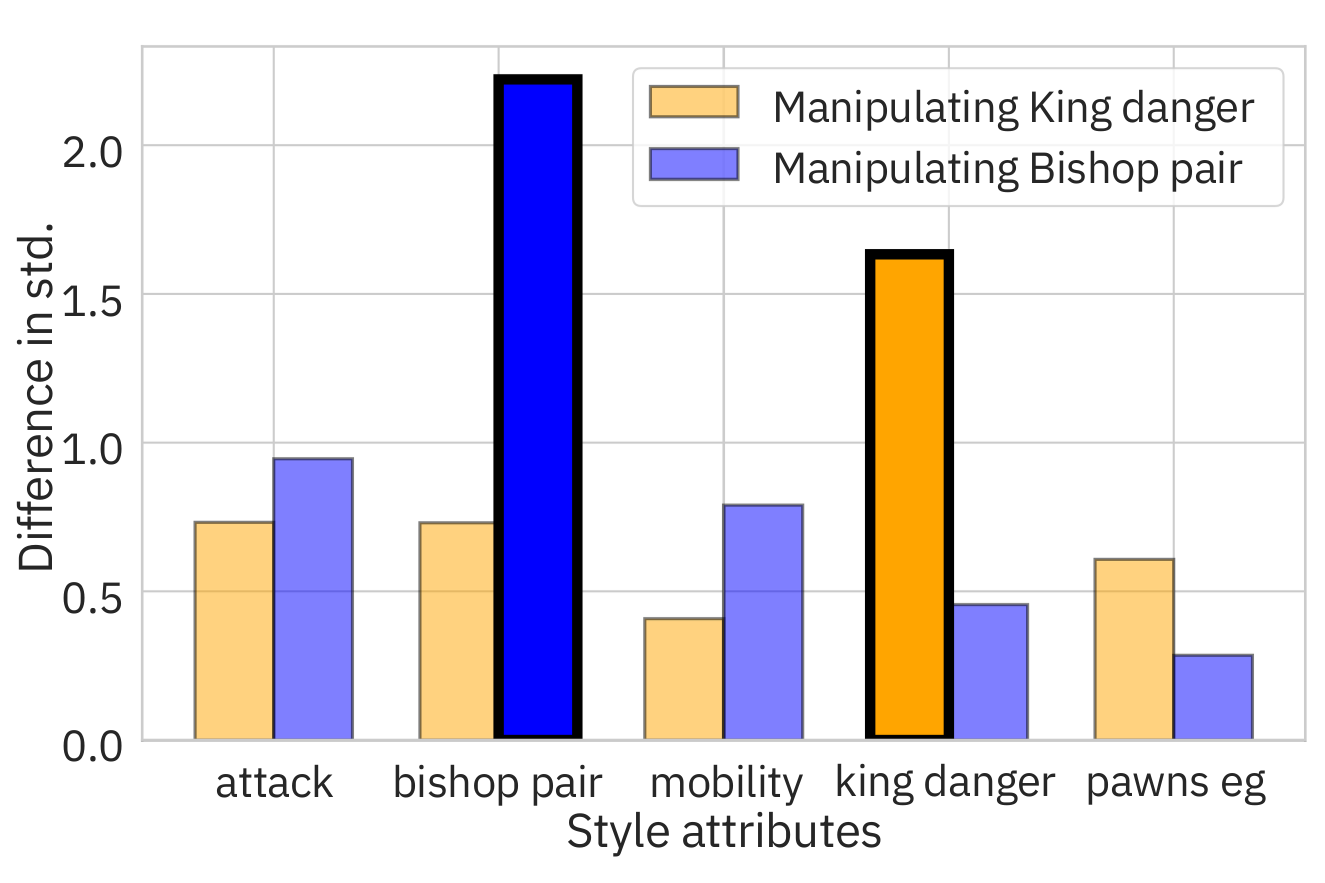}  
        \caption{Using our style steering method to increase two Stockfish attributes (separately) for 2,000 random players. }  
        \label{fig:steered}  
    \end{minipage}  
\end{figure}  

\paragraph{Interpolating between players.} 
We show that interpolating between the style vectors of a weaker and stronger player results in new players whose skill levels also interpolates between the players. %
Here, we take 100 pairs of weak and strong player style vectors and gradually interpolate between them as $(1 - \lambda) u_w + \lambda u_s$, $0 \leq \lambda \leq 1$, where $u_w$ and $u_s$ are the respective vectors. For each value of $\lambda$, we simulate 1,000 games between the interpolated player and $u_s$, the stronger player. 
Figure~\ref{fig:interpolation} plots the win rate of the interpolated players as a function of $\lambda$ for each pair of players. This plot shows that the win rate increases in a roughly linear fashion as lambda increases, starting low and eventually winning roughly half the time, which is what we would expect from two players with the same style vectors. This allows us to create a continuous range of skill levels, unlike current models such as \cite{mcilroy2020maia}.

\paragraph{Steering player style.} We can directly control the playing style of a player using the steering method described in Section~\ref{sec:style-methodology}. Using the human-interpretable Stockfish heuristics, we identify players in our chess dataset with high ($>2$ std) bishop pair utilization, and similarly players with high king danger. We use these player sets to compute style delta vectors corresponding to these attributes, and then simply add the delta vectors to 2,000 randomly sampled players' existing style vectors. Figure~\ref{fig:steered} shows the change (normalized by the standard deviation for that attribute) in these players' Stockfish evaluations after adding the style delta vectors. Indeed, we see that the player's style is steered towards the attribute in question, with modest impact on other attributes.

\begin{wrapfigure}[13]{R}{0.5\textwidth}
  \centering  
  \vspace{-16pt}
        \includegraphics[width=\linewidth]{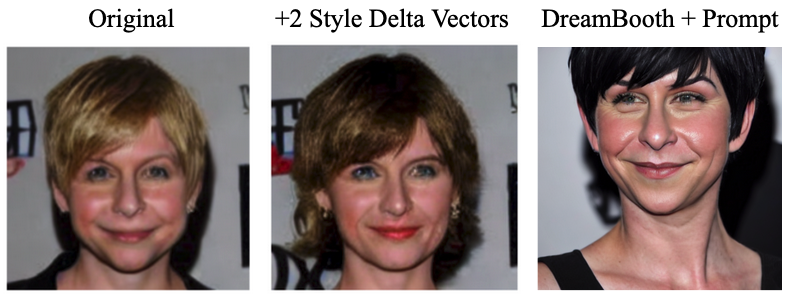}  
        \caption{Images generated by fine-tuning Stable Diffusion 1.5 with our methods on CelebA dataset. Our style steering to ``Black Hair" is compared to prompting DreamBooth.}  
        \label{fig:diffusion-steer-main}  

\end{wrapfigure}

\subsection{Application beyond gaming}
To show that our methods generalize beyond gaming applications, we apply the exact \polys fine-tuning method described in Section~\ref{subsec:traineval} and style steering method described in Section \ref{sec:style-methodology} to steer the generation of diffusion models. We fine-tune Stable Diffusion 1.5 \citep{stable15} on the CelebA Faces With Attributes dataset \citep{celeba} to create style vectors for the 10,177 identities. In Figure \ref{fig:diffusion-steer-main}, we compute a ``Black Hair" style delta vector using the cosine similarity between the images and their respective CLIP \citep{clip} embeddings, and use this to edit the image. To compare, we fine-tune Stable Diffusion 1.5 using DreamBooth \citep{dreambooth} using the scripts provided in \cite{peft} and edit the image by adding ``Black Hair" to the prompt. We provide more examples and comparisons in Appendix \ref{app:diffusion_steer} and Figure \ref{fig:diffusion_steer}.

\section{Other Related Work}

\paragraph{Stylometry and player style modeling.}

Originally referring to performing author attribution via statistical analysis of text \citep{tweedie1996neural, neal2017surveying}, stylometry has since come to refer to the general task of identifying individuals given a set of samples or actions, and has found broad application for tasks such as handwriting recognition \citep{bromley1993signature}, speaker verification \citep{wan2018generalized}, identifying programmers from code \citep{caliskan2015anonymizing}, determining user age and gender from blog posts \citep{goswami2009stylometric}, and identifying characteristics of authors of scientific articles \citep{bergsma2012stylometric}.
In the context of gaming (covered in the introduction),  %
stylometry is closely related to playstyle modeling, where the goal is to associate a player with a reference style, such as by building agents representative of different playstyles and find the closest behavioral match
\citep{holmgaard2014evolving}, or gathering gameplay data and applying methods such as clustering \citep{ingram2022play}, LDA \citep{gow2012unsupervised}, Bayesian approaches~\citep{normoyle2015bayesian}, and sequential models \citep{valls2015exploring} to identify groups of players with similar styles.
\citet{kanervisto2021generalcharacterizationagentsstates}
characterizes an agent's behavior by analyzing the states that an agent sees (not actions). Unlike our work, these approaches either focus on aggregate play styles, or do not learn generative models of behavior that can be conditioned on an individual's style.

Our method for style synthesis is inspired by earlier work on vector arithmetic with embeddings~\citep{church2017word2vec}, as well as recent work on steering multiask models with task vectors~\citep{ilharco2023editing}.
Our steering method is reminiscent of \citet{radford2016dcgan}, which manipulates the model's latent space to generate images containing specific attributes. Recently, \citet{dravid2024interpretingweightspacecustomized} achieved similar results on images of people by training LoRAs and manipulating their weights. 
\cite{acquisition-alphazero} probes chess concepts from a model to see if learned features are predictive of properties; we adopt their use of chess concepts when analyzing our player styles.

\paragraph{Parameter-efficient adaptation.}
 Approaches for efficient adaption of a pretrained model can be broadly grouped in two categories. Adapter based methods inject new parameters within a pretrained model, and only updates the newly inserted parameters while keeping the backbone fixed. \cite{houlsby2019parameter} defines an adapter as a two-layer feed-forward neural network with a bottleneck representation, and are inserted before the multi-head attention layer in Transformers. Similar approaches have been used for cross-lingual transfer \citep{pfeiffer-etal-2020-mad}. Adapters have also been used in vision based multitask settings \citep{rebuffi2017learning}. More recently, \cite{ansell-etal-2022-composable} propose to learn sparse masks, and show that these marks are composable, enabling zero-shot transfer. Lastly, \cite{lora} learn low-rank shifts on the original weights, and \cite{tfew} learns an elementwise multiplier of the pretrained model's activations. Adapters have also been used in multitask settings. \cite{chronopoulou-etal-2023-adaptersoup} independently trains adapters for each task, and merges parameters of relevant tasks to transfer to new ones. 

Another approach is the use of soft prompts \citep{lester-etal-2021-power}.
In the context of natural language, where the input is a sequence of work tokens, soft prompts appends learnable tokens to a natural language sequence. In a similar setting, \cite{vu2021spot} learns a collection of soft-prompts from a multitask training set, and given a novel task, retrieves relevant prompts for efficient transfer. 

\section{Conclusion}

We show that individual player behavior can be modeled at very large scale in games as different as chess and Rocket League. We cast this problem in the framework of multi-task learning and employ modular PEFT methods to learn a shared set of skills across players, modulated by distinct style vectors. 
We use style vectors to perform stylometry, analyze player styles, and synthesize and steer new styles. Our methods extend beyond gaming, as we show by using style vectors to edit the images of celebrities.

\bibliographystyle{abbrvnat}
\bibliography{arxiv}

\clearpage
\appendix
\section{Appendix}

\subsection{Multi-Head Adapter Routing} \label{app:mhr}
In \poly{}, the module combination step remains \textit{coarse}, as only linear combinations of the existing modules can be generated. \cite{caccia2022multi} propose a more fine-grained module combination approach, referred to as Multi-Head Routing (\polys{}). Similar to Multi-Head Attention \citep{transfo}, the input dimension of $\mA$ (and output dimensions of $\mB$) are partitioned into $h$ heads, where a \poly{}-style procedure occurs for each head. 
The resulting parameters from each head are then concatenated, recovering the full input (and output) dimensions. 
This makes the module combination step \textit{piecewise linear}, with a separate task-routing matrix $\mZ$ learned for each head. 

Formally, a \polys{} layer learns a 3-dimensional task-routing tensor $\tZ \in \mathbb{R}^{|\mathcal{T}| \times |\mathcal{M}| \times h}$. 
The 2D slice $\tZ_{:, :, k}\in\mathbb{R}^{|\mathcal{T}| \times |\mathcal{M}|}$ of the tensor $\tZ$ denotes the distribution over modules for the $k$-th head, and $\mW[k] \in \mathbb{R}^{\frac{d}{h} \times r}$ the $k$-th partition along the rows of the matrix $\mW \in \mathbb{R}^{d \times r}$. The adapter parameters $\mA^\tau \in \mathbb{R}^{d \times r}$ for task $\tau$, and for each adapter layer, are computed as (similarly for $\mB^\tau$):
\begin{gather*}
\tag{MHR}
\mA^\tau_{k} = \sum_{j} \alpha_{i,k} \cdot \mA_j[k]\;\; \textrm{with}\;\; \mA^\tau_{k} \in \mathbb{R}^{\frac{d}{h} \times r}, \\
 \mA^\tau = \texttt{concat}(\mA^\tau_{1}, \ldots, \mA^\tau_{h}), 
\label{eqn:polys}
\end{gather*}
where $\alpha_{i,k} = \texttt{softmax}(\mZ \texttt{[} \tau \texttt{,:,k]})_i$. Importantly, the number of LoRA adapter parameters does not increase with the number of heads. Only the task-routing parameters linearly increase with $h$ for \polys{} vs. \poly{}. However, this cost is negligible as the parameter count of the routing matrices is much smaller than for the LoRA modules themselves.

\subsection{Steering Diffusion Models}
\label{app:diffusion_steer}
To address questions about the generalizability of our method, we applied the exact style delta vector computation and steering algorithm outlined in Section~\ref{sec:style-methodology} to steer the outputs of an image generation diffusion model in a fine-grained manner. We use the CelebA Faces With Attributes dataset \citep{celeba} to fine tune style vectors for 10,177 identities. We use Stable Diffusion 1.5 \citep{stable15} as our base model. 

\begin{figure}
        \centering
        \includegraphics[width=0.8\linewidth]{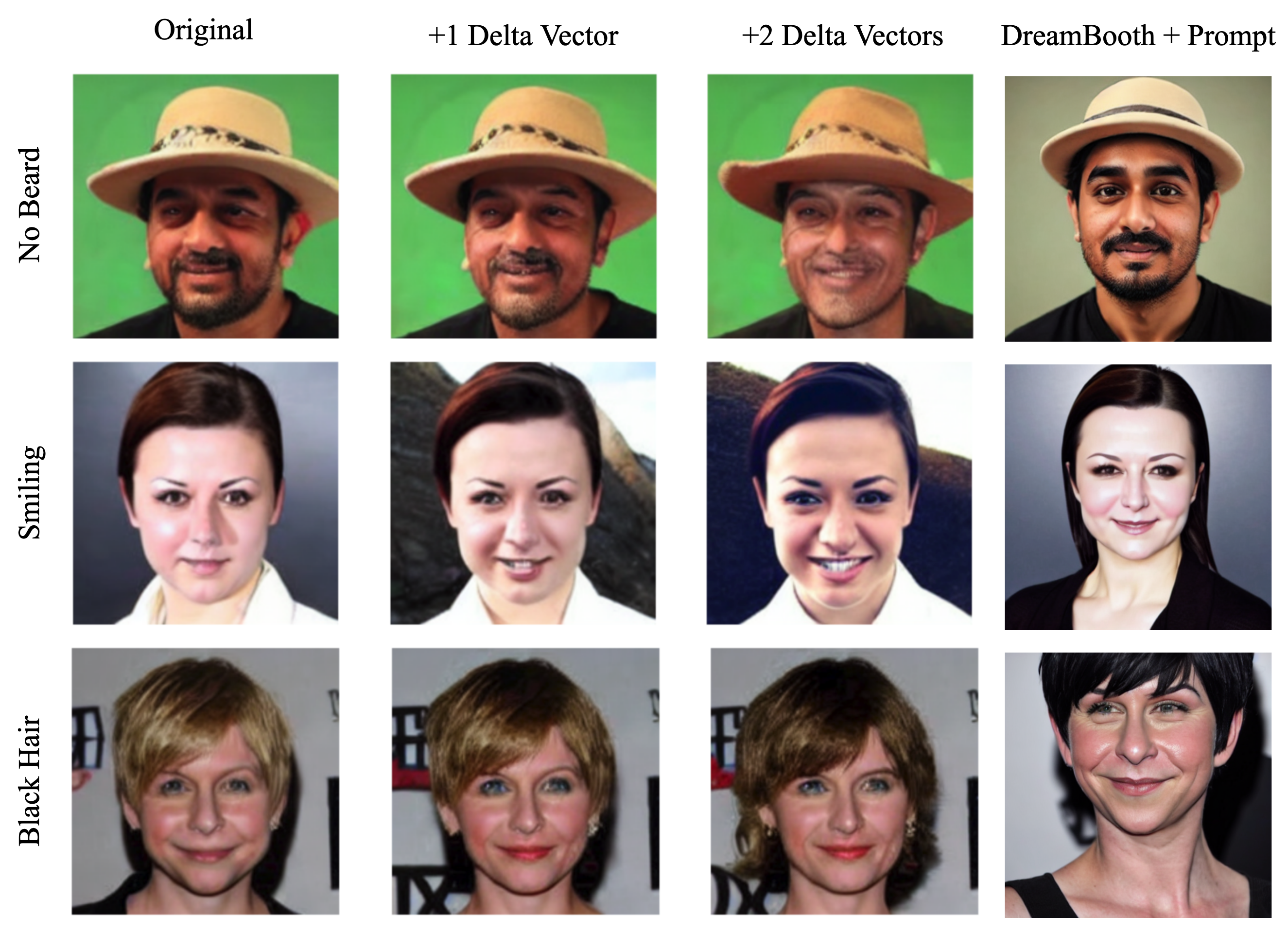}
        \caption{Images generated by steering Stable Diffusion 1.5 \citep{stable15} fine tuned with our method on the CelebA~\citep{celeba} dataset. We compare against using DreamBooth~\citep{dreambooth} on the original image and modifying the prompt.}
        \label{fig:diffusion_steer}
\end{figure}

We compute ``No Beard", ``Smiling", and ``Black Hair" style delta vectors using cosine similarities between the images and their respective CLIP \citep{clip} embeddings. Figure~\ref{fig:diffusion_steer} shows sample images generated by applying these vectors, with the leftmost images being un-steered. We compare our results against using DreamBooth~\citep{dreambooth} with LoRA using the scripts in \cite{peft} to fine-tune towards the original image, and adding "with no beard", "smiling", and "with black hair" to the prompt for the respective images. 

Our method is able to achieve more granular control of the source image with minimal modifications to the style of the image. In contrast, while DreamBooth is able to change the specific feature we aim to steer, the remaining parts of the image are changed significantly.

\subsection{Maia Architecture/Data}
\label{app:maia-details}

Our base Maia architecture follows \cite{mcilroy2022personalized} and uses  the Squeeze-and-Excitation (S\&E) Residual Network of \citep{hu2018squeeze}.
At every residual block, channel information is aggregated across spatial dimensions via a global pooling operation. The resulting vector is then processed by a 2-layer MLP, with a bottleneck representation compressing the number of channels by $r$. The output of this MLP is a one-dimensional vector used to scale the output of the residual block along the channel dimension. 
We use 12 residual blocks containing 256 filters, and a bottleneck compression factor of $r=8$. We note that this differs from the base Maia model in ~\cite{mcilroy2022personalized}, which uses 64 filters and 6 residual blocks.

\begin{figure}
    \centering
    \includegraphics[width=0.5\textwidth]{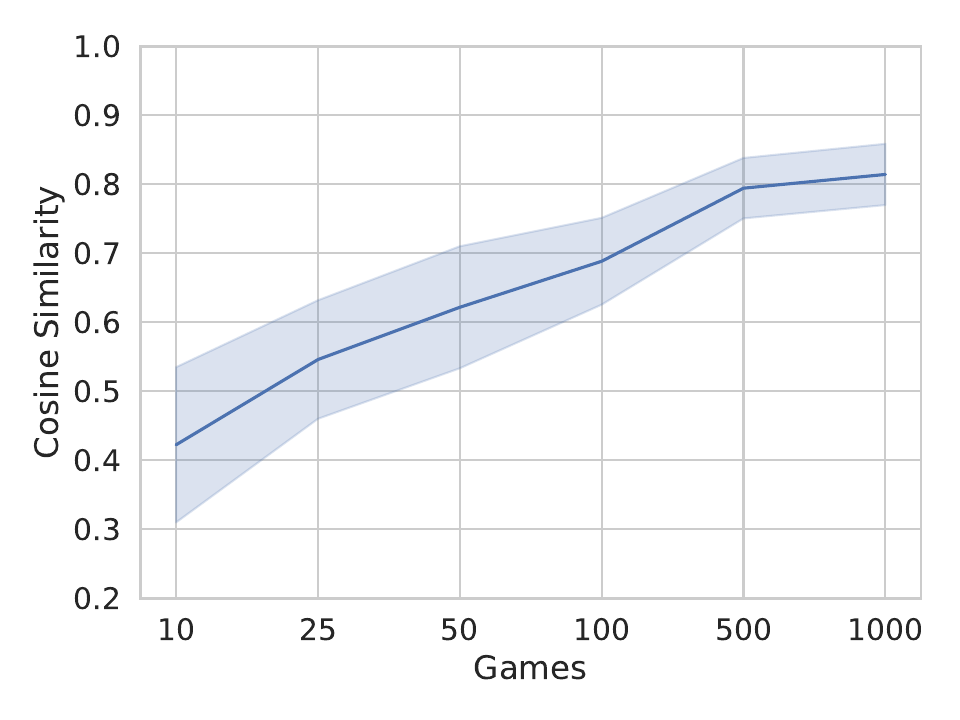}
    \caption{Cosine similarity of style vectors trained with varying game sizes compared to a style vector trained with 10,000 games, run on 50  players.}
    \label{fig:cosine_similarity_vargame}
\end{figure}

While our dataset has a median game count of 3,479 games, many players may have as few as 10-50 games, implying some degree of data imbalance. Our evaluation of few-shot learning shows that 100 games is sufficient to learn the style vector of an unseen player. However, one might still ask how accurately such a style vector is given a very small number of games. To explore this, we first split a player into disjoint sets of 10, 25, 50, 100, 500, and 1,000 games. We then train a style vector on each set. As a baseline, we train a style vector on 10,000 games and track the cosine similarity of the smaller-set style vectors relative to this baseline vector. We show the results in Figure~\ref{fig:cosine_similarity_vargame}. 

\subsection{Rocket League Architecture/Data}
\label{app:rocketleaguedata}

The 1v1 replays dataset was scraped over the course of several weeks from the Ballchasing.com API using the Grand Champion subscription tier, though the API does have a slower free tier. This API yields raw game replays, which are uploaded by users either manually or using a community-made plugin for the game. The replays are in a binary format which must be parsed using community-made projects such as Carball~\citep{carball}.

The Carball library allows us to convert the binary replay format to a more standard CSV format, which we save to a Cloud binary blob storage. The data present in both is a lossy reconstruction of game states, and requires some processing to be usable. In particular, the data is sampled at an inconsistent rate (varying between 24hz and 27hz), contains repeated physics ticks, and is missing action data for aerial controls (pitch, yaw, roll). 

We resolve the issue of sampling rate and repeated ticks by removing repeated ticks, and doing a time-weighted resampling and interpolation to a standard 10hz for model training, though we found that 30hz also works well. Note that the actual game physics ticks occur at 120hz, so any value aligned with this should work. Without these changes, the model performs extremely poorly and is unable to navigate the arena.

We resolve the issue of missing aerial controls through the physics-based solver present in the Carball library. The estimation of these controls is not perfect, but it is sufficient for our purposes. Some previous community work has used inverse dynamics~\citep{rolv} trained from rollouts of in-game bots to solve for these actions, though we opted to not use this due to the inconsistency in replay data sampling. 

The data returned by the CSVs are fairly large, messy, and inconsistent. We apply the following transformations to the dataframe to bring the values closer to 0:

\begin{itemize}
    \item Divide position by 2300 
    \item Divide linear velocity by 23000
    \item Divide angular velocity by 5500
    \item Divide boost by 255
    \item Encode rotation Euler angles according to \cite{zhou2020continuity}
\end{itemize}

Additionally, when turning the data into tokens for use in our model, we add in an extra dimension to represent the team, and concatenate the opponent's data points along with the position, linear and angular velocity of the ball. We complete all of these transformations at runtime.

We also have to align the data returned by the simulators for Rocket League with the data used to train the model, RLBot~\citep{rlbot} and RLGym~\citep{rlgym}. Along with including an extra dimension to represent the team, we apply the following transformations to all samples obtained from the game:

\begin{itemize}
    \item Divide position by 2300
    \item Divide linear velocity by 2300
    \item Divide angular velocity by 5.5
    \item Divide boost by 100
\end{itemize}

\begin{figure}
    \centering
    \includegraphics[width=0.5\textwidth]{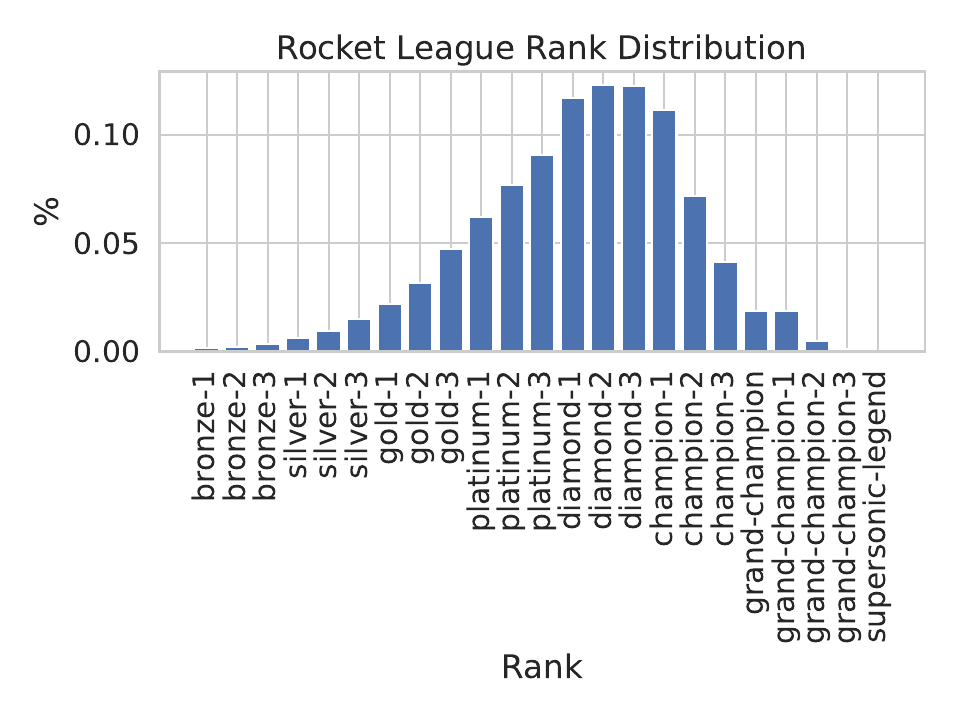}
    \caption{Skill distribution of Rocket League players in our dataset.}
    \label{fig:rocketleague_ranks}
\end{figure}

The skill distribution of the players in our dataset can be found in Figure~\ref{fig:rocketleague_ranks}. 

\begin{figure}
    \centering
    \includegraphics[width=0.5\textwidth]{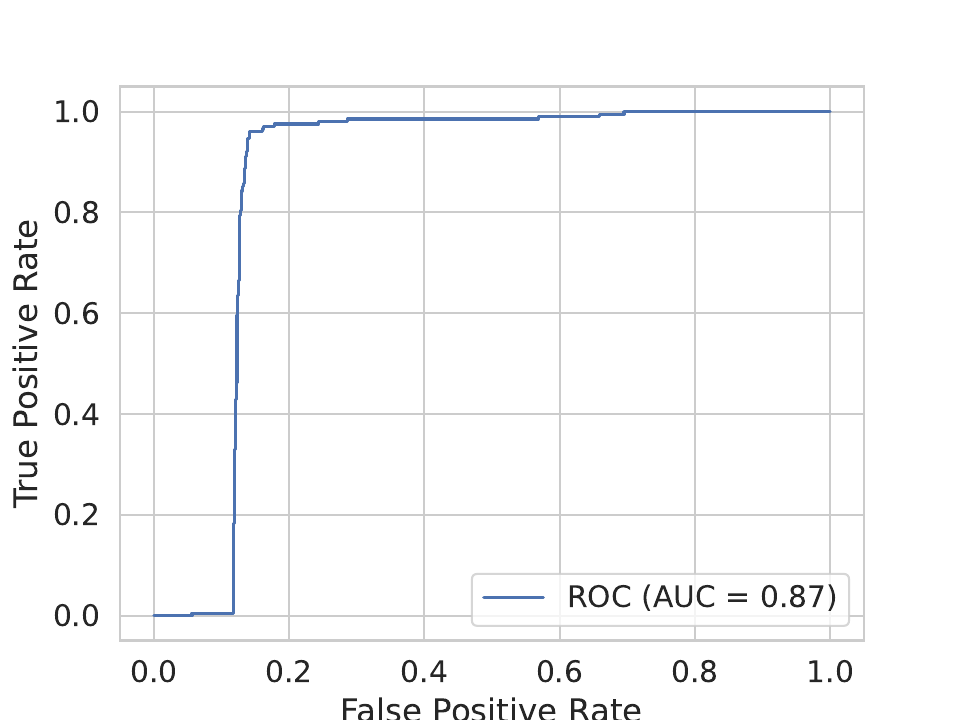}
    \caption{ROC Curve of Rocket League player detection.}
    \label{fig:roc_curve}
\end{figure}

\subsection{Implicit Stationarity Assumptions}
\label{app:assumptions}

Most of the existing work in chess assumes that a player remains stationary over time and across gameplay situations. However, in reality, a player's style may depend on the type of opponent they are facing, which opening is used, which stage of the game they are in (opening, middle, endgame), and so on. For instance, \cite{mcilroy2021stylometry} observe that stylometry accuracy drops when removing the opening (e.g., the first 15 moves) moves, suggesting that the opening has an outsized effect on style identification. Our approach does not rely on these assumptions and can in principle be applied to arbitrary subsets of a player's data. For instance, one could split a player's data into opening, middlegame, and endgame moves and train a separate style vector for each. One could further split the data based on which defense the opponent uses, what time of the day it is, etc.. Despite treating players holistically and avoiding any splits of their data, we are still able to capture the peculiarities of each individual's playing style and perform stylometry with high accuracy. This also enables us to compare our results to those of prior work, which also treats player data holistically.

\end{document}